\documentclass[sigconf]{acmart}

\AtBeginDocument{%
  }
\usepackage{multirow}
\usepackage{arydshln}  %
\usepackage{array} 
\usepackage[utf8]{inputenc}
\usepackage{url}
\usepackage{booktabs}
\usepackage{makecell}
\usepackage{arydshln}
\usepackage{bbding}
\usepackage{pifont}
\usepackage{wasysym}
\usepackage{utfsym}
\usepackage{fontawesome}
\usepackage[table]{xcolor} 
\usepackage{booktabs}      
\usepackage{pifont}        
\usepackage{booktabs}     
\usepackage{tabularx}     
\usepackage{graphicx}     
\copyrightyear{2025}
\acmYear{2025}
\setcopyright{acmlicensed}\acmConference[MM '25]{Proceedings of the 33rd ACM International Conference on Multimedia}{October 27--31, 2025}{Dublin, Ireland}
\acmBooktitle{Proceedings of the 33rd ACM International Conference on Multimedia (MM '25), October 27--31, 2025, Dublin, Ireland} \acmDOI{10.1145/3746027.3755174} \acmISBN{979-8-4007-2035-2/2025/10}
\begin{document}

\title{CookAnything: A Framework for Flexible and Consistent Multi-Step Recipe Image Generation}

\author{Ruoxuan Zhang}
\email{zhangrx22@mails.jlu.edu.cn}
\affiliation{%
  \institution{Jilin University}
  \city{Changchun}
  \country{China}
}

\author{Bin Wen}
\email{wenbin2122@mails.jlu.edu.cn}
\affiliation{%
  \institution{Jilin University}
  \city{Changchun}
  \country{China}
}

\author{Hongxia Xie}
\authornote{Corresponding Author.}
\email{hongxiaxie@jlu.edu.cn}
\affiliation{%
  \institution{Jilin University}
  \city{Changchun}
  \country{China}
}

\author{Yi Yao}
\email{leo81005.ee10@nycu.edu.tw}
\affiliation{%
  \institution{National Yang Ming Chiao Tung University}
  \city{Hsinchu}
  \country{Taiwan}
}

\author{Songhan Zuo}
\email{zuosh2122@mails.jlu.edu.cn}
\affiliation{%
  \institution{Jilin University}
  \city{Changchun}
  \country{China}
}

\author{Jian-Yu Jiang-Lin}
\email{jianyu@cmlab.csie.ntu.edu.tw}
\affiliation{%
  \institution{National Taiwan University}
  \city{Taipei}
  \country{Taiwan}
}

\author{Hong-Han Shuai}
\email{hhshuai@nycu.edu.tw}
\affiliation{%
  \institution{National Yang Ming Chiao Tung University}
  \city{Hsinchu}
  \country{Taiwan}
}

\author{Wen-Huang Cheng}
\email{wenhuang@csie.ntu.edu.tw}
\affiliation{%
  \institution{National Taiwan University}
  \city{Taipei}
  \country{Taiwan}
}
\renewcommand{\shortauthors}{Ruoxuan Zhang et al.}


\begin{abstract}
Cooking is a sequential and visually grounded activity, where each step such as chopping, mixing, or frying carries both procedural logic and visual semantics. While recent diffusion models have shown strong capabilities in text-to-image generation, they struggle to handle structured multi-step scenarios like recipe illustration. Additionally, current recipe illustration methods are unable to adjust to the natural variability in recipe length, generating a fixed number of images regardless of the actual instructions structure.
To address these limitations, we present CookAnything, a flexible and consistent diffusion-based framework that generates coherent, semantically distinct image sequences from textual cooking instructions of arbitrary length. The framework introduces three key components: (1) Step-wise Regional Control (SRC), which aligns textual steps with corresponding image regions within a single denoising process; (2) Flexible RoPE, a step-aware positional encoding mechanism that enhances both temporal coherence and spatial diversity; and (3) Cross-Step Consistency Control (CSCC), which maintains fine-grained ingredient consistency across steps.
Experimental results on recipe illustration benchmarks show that CookAnything performs better than existing methods in training-based and training-free settings. The proposed framework supports scalable, high-quality visual synthesis of complex multi-step instructions and holds significant potential for broad applications in instructional media, and procedural content creation. More details are at \url{https://github.com/zhangdaxia22/CookAnything}.

\end{abstract}


\begin{teaserfigure}
  \includegraphics[width=\textwidth]{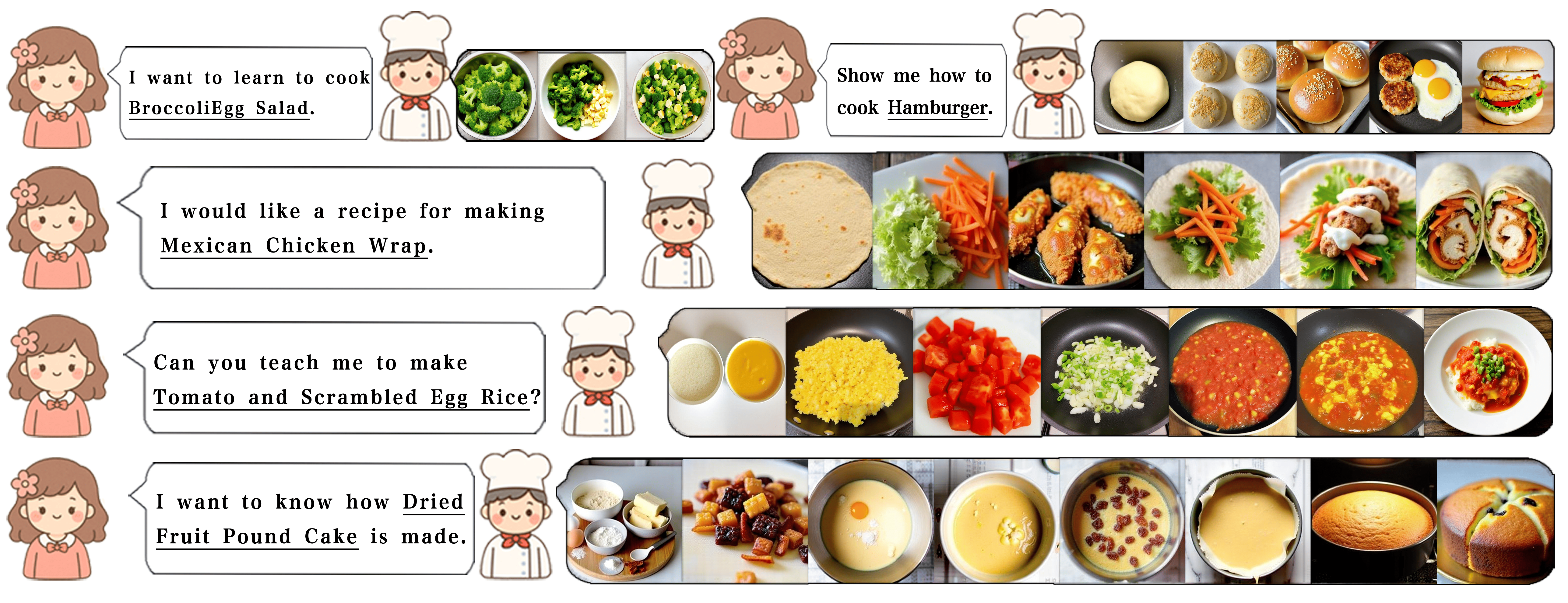}
   \caption{Demonstration of our CookAnything model generating multi-step cooking instructions in a single pass. Each example shows the user’s prompt (left) and the corresponding series of dish images (right), from initial preparation steps through the final plated result (Details of the complete recipe text can be found in the Supplementary A.6.).}
  \label{fig:teaser}
\end{teaserfigure}


\maketitle
\section{Introduction}

Cooking is a richly visual and sequential activity: from chopping onions to garnishing a dish, each step not only involves semantic transitions but also yields observable visual transformations~\cite{semantic1, visual1,visual2}. Accurately illustrating these processes from textual instructions holds significant value for applications in culinary education, assistive technology, and multimodal content generation, enabling users to better understand, follow, and interact with complex procedures in an intuitive visual manner.

As textual recipes abstract the cooking process into language, recipe illustration aspires to reverse this abstraction, generating coherent image sequences that visually narrate each procedural step~\cite{menon2024generating,han2020cookgan, liu2023ml, pan2020chefgan}. Compared to single-image generation, this task introduces unique challenges: \textit{it requires maintaining temporal progression, preserving ingredient consistency, and capturing subtle visual distinctions between stages}. 
StackedDiffusion~\cite{menon2024generating} pioneered the task of illustrated recipe instructions by generating one image per recipe step. However, its design assumes a \textit{fixed number of steps}, ignoring the inherent variability across recipes, resulting in both under-generation and over-generation in real-world settings.

While recent advances in text-to-image synthesis, particularly diffusion-based models such as FLUX~\cite{flux1ai2024} have achieved remarkable success in high-fidelity image generation, these models are predominantly designed for single-image outputs. This limits their applicability in structured, multi-step domains such as cooking recipes, where each visual output must correspond to a distinct semantic step and together form a coherent sequence. Attempts to extend these models, such as In-Context LoRA~\cite{IC-Lora}, adopt a simple concatenation of step prompts to jointly synthesize multi-step outputs. However, this design leads to \textit{semantic entanglement, where visual features bleed across steps, producing indistinguishable or incoherent images that undermine the narrative flow}.

Motivated by these challenges, we investigate the key question: \textbf{How can we achieve flexible, coherent, and semantically disentangled multi-step recipe image generation in a unified framework?}

We propose \textbf{CookAnything}, a diffusion-based framework for generating step-by-step illustrated recipes with variable length and high visual consistency (as shown in Fig.~\ref{fig:teaser}). It introduces three key components: (1) a \textbf{Step-wise Regional Control (SRC)} mechanism that assigns each instruction to a distinct latent region, ensuring semantic separation and global coherence; (2) a new positional encoding method, \textbf{Flexible Rotary Position Embedding (RoPE)}, which resets coordinate indices per step to support diverse layouts; and (3) a \textbf{Cross-Step Consistency Control (CSCC)} module that preserves the visual continuity of fine-grained ingredients across steps. These innovations enable structured, coherent, and flexible multi-image generation for sequential visual synthesis. 

Our contributions are summarized as follows:
\begin{itemize}
    \item We propose \textbf{CookAnything}, the first diffusion-based framework for illustrated recipe generation with arbitrary-length, step-wise image sequences, flexibly adapting to diverse real-world structures.
    \item We introduce \textbf{SRC} and \textbf{Flexible RoPE} to address position misalignment via step-aware spatial encoding and region binding, and \textbf{CSCC} to ensure visual consistency of recurring ingredients across steps.
    \item Experiments show that our method achieves \textit{state-of-the-art} results in both training-based and training-free settings, with broad potential in instructional and procedural content generation.
\end{itemize}

\section{Related Work}

\subsection{Recipe Analysis Task}

The study of food-centric multimedia has garnered increasing attention within the multimedia research community due to its significant relevance to human survival, nutrition, health, and sensory enjoyment~\cite{yin2023foodlmm, zhou2024foodsky, mm-recipe-1,mm-recipe-2,mm-recipe-dataset,mm-recipe3,mm-recipe4,mm-recipe5}. The growing availability of large-scale food datasets such as Recipe1M+ \cite{marin2021recipe1m+}, Food-101 \cite{bossard2014food}, VireoFood-172 \cite{chen2016deep}, and Nutrition5k \cite{thames2021nutrition5k} has fueled research in a wide range of fine-grained food analysis tasks.
These tasks include food and ingredient classification \cite{min2023large,8779586,min2020isia,zhang2024recognize,zhang2025learning}, food instance segmentation \cite{lan2023foodsam,wu2021large}, and nutrition or weight estimation \cite{gui2024navigating}, among others. 

In this work, we focus on the task of step-wise recipe image generation, which aims to synthesize a sequence of visual illustrations corresponding to each step in a cooking recipe.
Early efforts, such as CookGAN \cite{han2020cookgan}, generated dish images based on latent representations of ingredient lists, while ChefGAN \cite{pan2020chefgan} employed recipe instructions as input. ML-CookGAN \cite{liu2023ml} further combined both ingredients and steps to generate the final dish image. 
However, these methods are inherently limited to producing only a single image corresponding to the completed dish, thus failing to capture the procedural nature of cooking.
To address this limitation, StackedDiffusion \cite{menon2024generating} introduced the novel task of illustrated recipe instructions, where an image is generated for each individual step in the recipe. However, it falls short in adapting to the natural variability of recipe lengths, producing a fixed number of step images regardless of the actual recipe structure.
In this work, we propose a flexible framework, \textbf{CookAnything}, that dynamically adapts to the varying number of steps in different recipes while ensuring accurate visual-semantic alignment, procedural coherence, and ingredient consistency across the generated sequence.

\subsection{Procedural Sequence Generation Model} 
In the context of text-to-image synthesis, a Procedural Sequence Generation Model decomposes the generation pipeline into discrete stages or operations, such as layout planning, object placement, attribute assignment, and appearance refinement.
Recently, diffusion models have revolutionized text-to-image generation, producing high-fidelity visuals through iterative denoising \cite{flux1ai2024, IC-Lora, mm-diffusion1, mm-diffusion2}. Latent Diffusion Models (LDM) \cite{LDM} and Vision Transformer-based Diffusion Transformers (DiTs) \cite{DiT, flux1ai2024} balance generation quality with scalability, capturing global semantics via attention mechanisms. However, most existing work focuses on single-image synthesis, overlooking structured, multi-step image generation.

Emerging efforts in story visualization \cite{maharana2022storydall} hint at the potential of sequential visual generation but fall short in domains like recipes, which demand procedural consistency, spatial coherence, and semantic disentanglement across steps. Crucially, ingredient continuity must be preserved, ensuring logical visual evolution throughout the cooking process.

To our knowledge, we are the first to introduce a diffusion-based model for step-wise recipe illustration. Building on the DiT backbone, our framework explicitly models semantic grounding, temporal alignment, and ingredient consistency across visual sequences, offering a structured and extensible solution for procedural image generation.

\section{Method}

In this section, we introduce \textbf{CookAnything},  a framework for flexible and consistent multi-step recipe image generation (see Fig.~\ref{fig:model}). 
To enable the generation of multiple step-specific images in a single denoising process, we propose a \textbf{Step-wise Regional Control (SRC)} mechanism, which aligns each step instruction with its corresponding visual region. However, standard Rotary Positional Encoding (RoPE) suffers from positional misalignment and weakened long-range dependencies when applied across multiple regions. 
Additionally, to tackle the \textbf{Tiny Ingredient Continuity Problem} where small but critical ingredients may be visually inconsistent or even missing across steps, we design \textbf{Cross-Step Consistency Control (CSCC)}. CSCC utilizes Contextual Step Tokens to promote visual consistency of shared ingredients without compromising step-wise independence.

\subsection{Preliminary}

\begin{figure*}[t]
  \centering
  \includegraphics[width=1\linewidth]{arc.png} 
  \caption{Overall structure of our \textit{CookAnything} model, illustrated with a 3-step vegetable pancake recipe. The Cooking Agent reformats the raw recipe into context-tagged steps, supplementing missing ingredient details. Each step is encoded by a \textit{T5 Encoder }in two ways: (1) all steps are concatenated to capture global context and produce \textit{contextual step tokens}, and (2) each step is encoded independently to preserve local semantics and generate \textbf{step tokens}. These two types of tokens are fused via\textbf{ weighted averaging}. Meanwhile, noisy latent tokens, processed by Flexible RoPE, are fed into DiT. A \textit{Step-wise Regional Attention Mask} is applied during DiT’s self-attention to constrain attention within each step, ensuring step-wise focus and visual consistency. In the illustration, purple, green, and pink tokens represent Steps 1, 2, and 3, respectively}
  \label{fig:model}
\end{figure*}

\textbf{Flux.1-dev.}
Flux.1-dev is a text-to-image model that generates a single high-quality image from a text prompt~\cite{flux1ai2024}. It replaces the U-Net \cite{unet} in Stable Diffusion (SD) \cite{SD} with a Diffusion Transformer (DiT)\cite{DiT} for better representation learning. For text encoding, it combines T5~\cite{T5} and CLIP~\cite{radford2021learning} to improve text-image alignment. DiT performs joint self-attention over concatenated text and latent tokens, processing noisy latent tokens $\mathbf{z} \in \mathbb{R}^{N \times d}$ and text condition tokens $\mathbf{C}^T \in \mathbb{R}^{M \times d}$, where $d$ is the embedding dimension, and $N$, $M$ are the numbers of image and text tokens.

\noindent\textbf{RoPE.}
In the Flux.1-dev model, Rotary Position Embedding (RoPE) is employed to encode positional information for latent tokens. Given noisy latent tokens $z$, the positional encoding process can be mathematically expressed as:
 \begin{equation}
     \dot{z}_{i,j} = z_{i,j} \cdot R(i,j),
 \end{equation}
 
\noindent where $R(i,j)$ is the rotation matrix corresponding to the position $(i,j)$, effectively encoding the spatial location of each token within the image. This approach enhances the model's ability to capture spatial relationships and dependencies inherent in visual data.

\noindent\textbf{Joint Attention.}
The joint attention mechanism maps position-encoded tokens into Query $Q$, Key $K$, and Value $V$. Additionally, it concatenates the text tokens for attention calculation. The attention operation can be expressed as:
\begin{equation}
    \text{Attn}([C^T;\dot{z}_{i,j} ]) = \text{SoftMax}\left( \frac{Q K^\top}{\sqrt{d}} \right) V,
\end{equation}
where $Q$, $K$, and $V$ are the queries, keys, and values derived from the token embeddings, and 
$d$ represents the dimensionality of the embeddings. The concatenation of the image and text tokens, denoted as $[ C^T;\dot{z}_{i,j} ]$, facilitates the multi-modal attention mechanism, enabling joint attention across both modalities.

\subsection{Step-wise Regional Control}

\noindent\textbf{Limitations of Flux.1-dev.} 
Flux.1-dev is designed for generating \textit{a single image} from a text prompt, making it inherently unsuitable for tasks that require \textit{a coherent sequence of step-wise images}—such as recipe visualization. This application demands accurate per-step generation, global consistency, and visual diversity across steps. To adapt Flux.1-dev for multi-image generation, In-Context LoRA~\cite{IC-Lora} proposes a simple strategy: concatenating all step-level prompts and jointly generating the full sequence of images in a single pass. While this method allows simultaneous generation, it \textbf{lacks explicit step-level separation}, resulting in severe \textit{semantic entanglement} across steps. Visual features from one instruction often leak into others, producing \textbf{highly similar and indistinct images} across steps.

We quantify this limitation using the Cross-Step Consistency (CSC) metric. The In-Context LoRA baseline yields a CSC score of 44.12—\textbf{9.03 points lower} than the ground truth score of 53.15—and the \textbf{lowest among all evaluated methods} (more details can be found in Tab.~\ref{tab:maintab}). This significant drop clearly indicates that \textit{without explicit step-level control, the model fails to maintain step-wise distinctiveness and coherence}.

\noindent\textbf{Step-wise Regional Control (SRC).} 
To address the entanglement and lack of step-wise control in Flux.1-dev and its simple concatenation-based variants, we propose SRC-a novel mechanism that enables the model to synthesize a coherent sequence of semantically distinct step images \textit{within a single denoising process}. SRC introduces architectural changes that explicitly bind each textual step instruction to a designated image region while preserving global coherence across the entire image. This allows for both localized control and smooth visual transitions, bridging the gap between single-image generation and structured multi-image synthesis.

\noindent\textbf{Step-wise Encoding and Integration.}
SRC modifies the decoding pipeline of Flux.1-dev by introducing a \textbf{Step-wise Encoding Mechanism} and a \textbf{Step-wise Regional Attention Mask}. Specifically, each recipe step is first independently encoded using a shared text encoder, and the resulting step tokens are concatenated before the latent tokens:
\begin{equation}
    C^{\text{input}} = \left[C^{(1)}; C^{(2)}; \dots; C^{(N)} \right],
\end{equation}
\begin{equation}
    X^{\text{input}} = \left[ C^{\text{input}}; z^{\text{input}} \right],
\end{equation}
where $C^{(n)}$ denotes the encoded tokens of the $n$-th step and $z^{\text{input}}$ denotes the noisy latent tokens.

\noindent\textbf{Step-wise Regional Attention Mask.}
To prevent semantic leakage between steps and ensure localized step-to-region alignment, we design a \textbf{Step-wise Regional Attention Mask} $M \in \mathbb{R}^{2N \times 2N}$ restricting attention within each step-region pair. Formally, it is defined as:
\begin{equation}
    M^{i,j} = 
    \begin{cases} 
    1 & \text{if } i = j \text{ or } |i-j|=N, \\
    0 & \text{otherwise}.
    \end{cases}
\end{equation}
Here, $i$ and $j$ represent the $i$-th and $j$-th step/image token sets. The attention operation within DiT then becomes:
\begin{equation}
    \text{Attn}(Q, K, V, M) = \text{Softmax} \left( \frac{Q K^T}{\sqrt{d_k}} \odot M \right) V,
\end{equation}
ensuring each step attends only to its paired visual region and associated instruction tokens.

The updated regional latent representation at timestep $t-1$ is computed as:
\begin{equation}
    z_{\text{region}}^{t-1} = \psi\left(\text{Attn}(Q_{\text{region}}^{t-1}, K_{\text{region}}^{t-1}, V_{\text{region}}^{t-1}, M_{\text{region}})\right),
\end{equation}
where $\psi$ represents the DiT block from Flux.1-dev.

\textbf{Whole-Description Control for Global Coherence.}
To complement regional specificity with global visual consistency, we incorporate a \textbf{Whole-Description Control Mechanism} that processes the full recipe description in parallel:
\begin{equation}
    z_{\text{base}}^{t-1} = \psi(\text{Attn}(Q_{\text{base}}^{t-1}, K_{\text{base}}^{t-1}, V_{\text{base}}^{t-1}, M_{\text{base}})).
\end{equation}

Finally, we fuse the regional and global latent representations through a weighted interpolation:
\begin{equation}
    z_{t} = \alpha \cdot z_{t-1}^{\text{base}} + (1-\alpha) \cdot z_{t-1}^{\text{region}},
\end{equation}
where $\alpha \in [0,1]$ controls the trade-off between global context and regional detail.

By preserving global structure while controlling step-wise semantics and region-level generation, SRC overcomes the limitations of prior methods and enables high-quality procedural image synthesis.

\setlength{\floatsep}{5pt plus 2pt minus 2pt}
\setlength{\textfloatsep}{5pt plus 2pt minus 2pt}
\setlength{\intextsep}{5pt plus 2pt minus 2pt}
\subsection{Flexible RoPE}
\begin{figure}[t]
  \centering
 
  \includegraphics[width=1\linewidth]{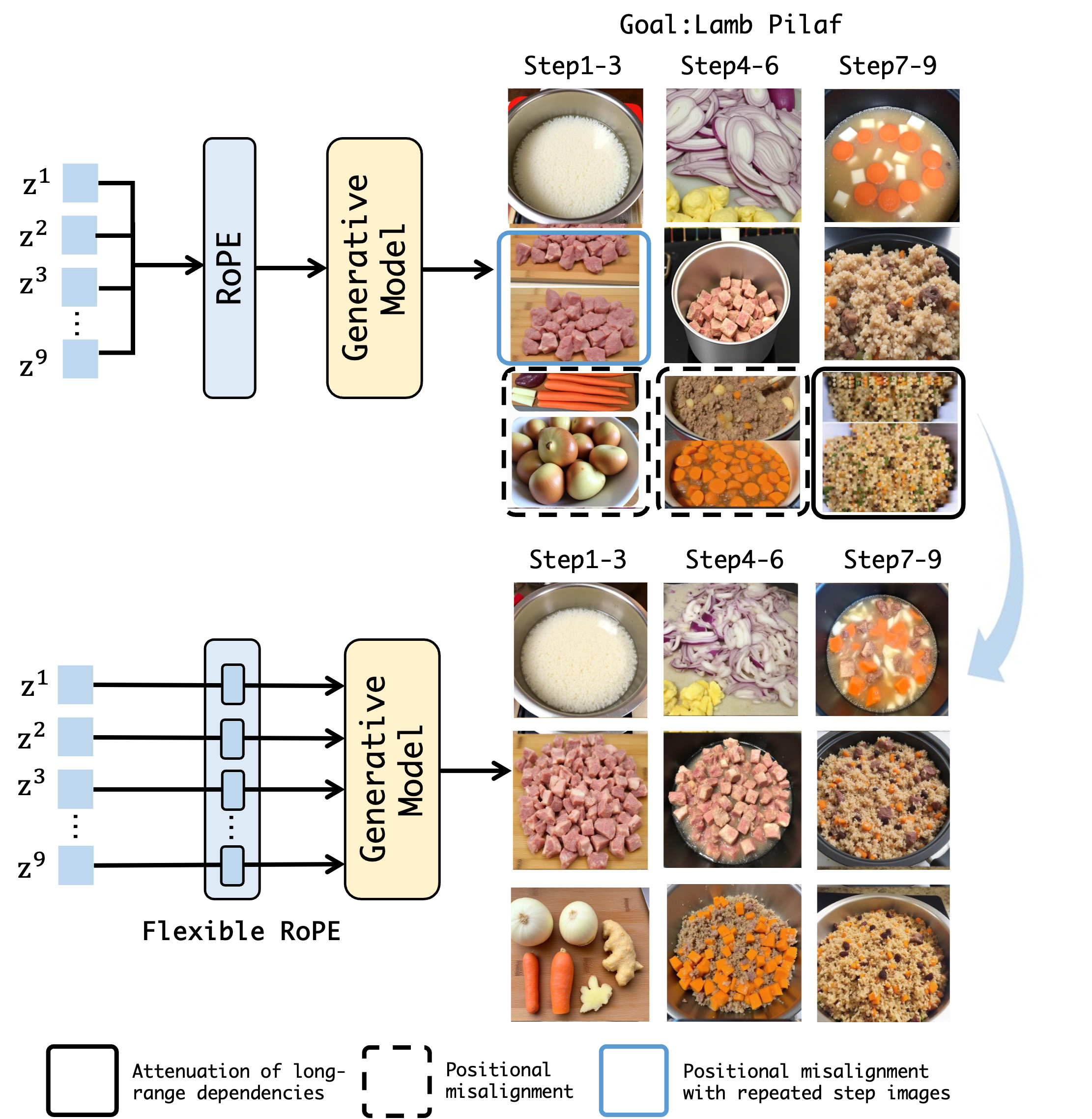} 

  \caption{The example from Original RoPE. Visualization comparison between original RoPE and our proposed Flexible RoPE using the example of \textit{Lamb Pilaf}. With original RoPE, repeated step images appear as early as Step 2. Steps 3 and 6 exhibit positional misalignment, and Step 9 suffers from noticeable blurring. In contrast, Flexible RoPE maintains clear step-wise differentiation, stable spatial alignment, and improved visual sharpness throughout the cooking process.}
  \label{fig:rope}
\end{figure}
\setlength{\floatsep}{5pt plus 2pt minus 2pt}
\setlength{\textfloatsep}{5pt plus 2pt minus 2pt}
\setlength{\intextsep}{5pt plus 2pt minus 2pt}

\textbf{Limitation of the Origin RoPE.} When generating a flexible number of step-wise images, Flux.1-dev, which adopts the original Rotary Position Embedding (RoPE), encounters two key limitations, as illustrated in Fig.~\ref{fig:rope} using a \textit{Lamb Pilaf} example.
The first issue is the \textbf{Misaligned Positional Embedding}. When generating multi-images in a single forward pass using FLUX, the use of original RoPE causes the positional encoding across steps to remain entangled in a global coordinate frame.  RoPE tends to overemphasize absolute positional alignment, which is suboptimal for tasks requiring local step-wise independence. As a result, each step image in our generation process is decoded from a similar latent spatial origin, leading to visual redundancy and layout collapse. In Fig.~\ref{fig:rope}, Step 2 is visually repeated, and the semantic boundary between steps becomes ambiguous.
The second issue is the \textbf{ Attenuation of Long-range Dependencies}. As step count grows, the model struggles to maintain semantic consistency, causing blurry or collapsed outputs in later steps—clearly seen at Step 9 in Fig.~\ref{fig:rope}.

To address this, we propose a step-aware positional encoding that explicitly re-initializes positional indices for each step, enabling the model to better capture both step-wise independence and inter-step coherence.

\noindent\textbf{Flexible RoPE.}
Unlike standard RoPE, which uses a globally continuous encoding across all image regions and leads to entangled positional dependencies, our proposed \textbf{Flexible RoPE assigns an independent positional encoding to each image region}. This disentanglement allows the model to clearly differentiate the position and semantics of each step, reducing cross-region interference and preserving generation fidelity even when scaling to many steps.

Specifically, for each image region $n$, we apply a separate RoPE encoding: 
\begin{equation} \dot{z}_{i,j}^{(n)} = z_{i,j}^{(n)} \cdot R^{(n)}(i,j), \end{equation} 
where $R^{(n)}(i,j)$ is the region-specific rotation matrix for the $n$-th image, applied to token $z_{i,j}^{(n)}$ at position $(i,j)$. This design ensures that each region learns positionally independent patterns, effectively preventing the model from inheriting noise or structural artifacts from adjacent steps.

We then concatenate all positionally encoded tokens as input to the model: 
\begin{equation} z_{\text{input}} = \left[ \dot{z}^{(1)}; \dot{z}^{(2)}; \dots;\dot{z}^{(N)} \right], \end{equation}
where $N$ denotes the number of step-wise images. This forms a joint input sequence where each region maintains its own positional integrity while enabling global attention across regions.

In Fig.\ref{fig:rope}, Flexible RoPE leads to significantly improved spatial alignment and visual consistency across steps, especially in long recipes, demonstrating its effectiveness and scalability.

\setlength{\floatsep}{5pt plus 2pt minus 2pt}
\setlength{\textfloatsep}{5pt plus 2pt minus 2pt}
\setlength{\intextsep}{5pt plus 2pt minus 2pt}
\subsection{Cross-Step Consistency Control}
\begin{figure}[t]
  \centering
  \includegraphics[width=1\linewidth]{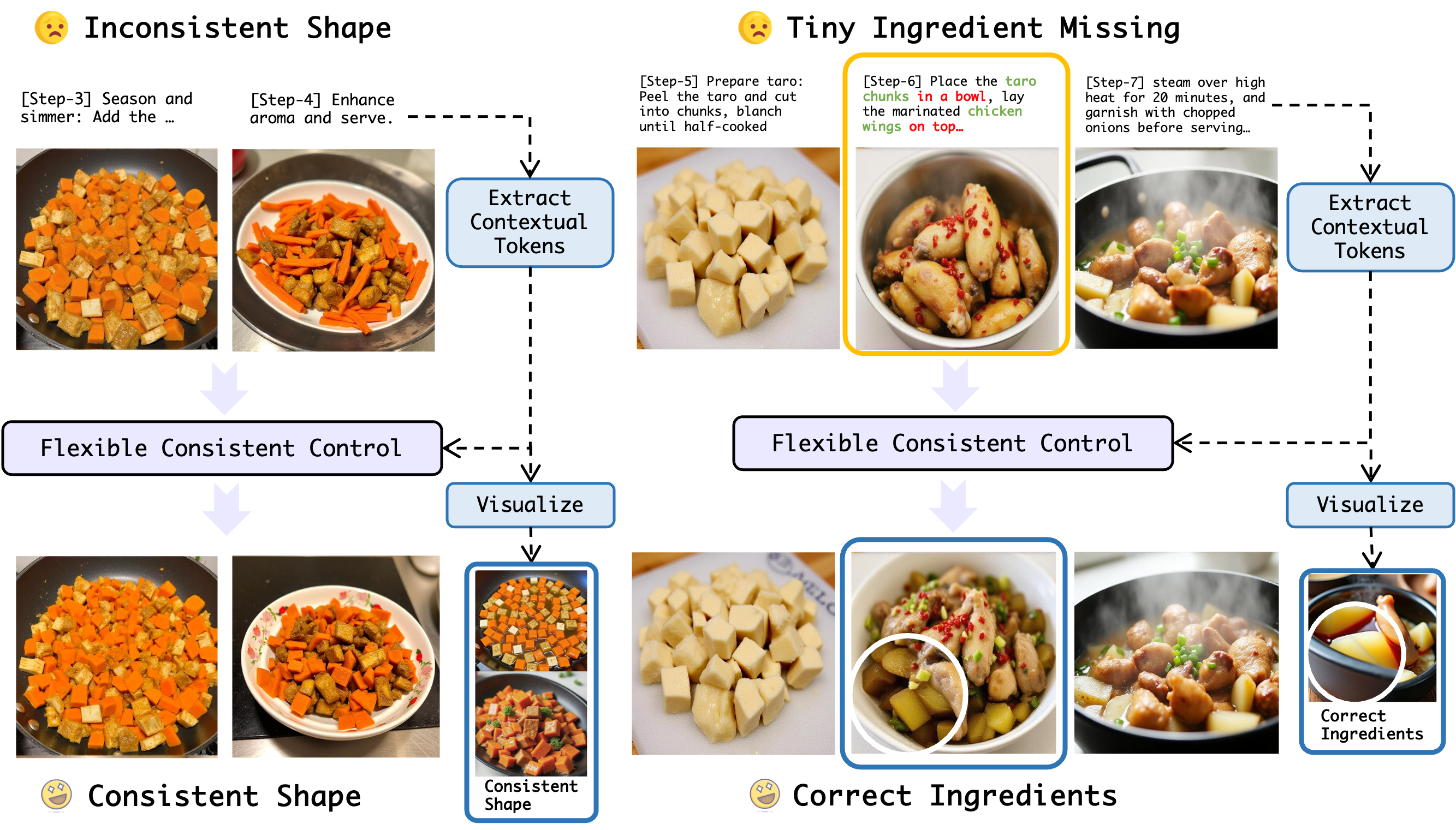} 
  \caption{Examples before and after applying Cross-Step Consistency Control (CSCC). \textit{Left: Stir-Fried Carrot with Dried Tofu.} Without CSCC, the carrot changes from cubes to strips in Step 4. Visualization of contextual tokens (using Flux.1-dev) shows shape continuity is preserved, so CSCC helps maintain a consistent appearance.
 \textit{Right: Steamed Chicken Wings with Taro.} In Step 5, taro should appear beneath the wings but disappears without CSCC. Since contextual tokens confirm its presence, CSCC successfully preserves it.}
  \label{fig:CSCC}
\end{figure}
\setlength{\floatsep}{5pt plus 2pt minus 2pt}
\setlength{\textfloatsep}{5pt plus 2pt minus 2pt}
\setlength{\intextsep}{5pt plus 2pt minus 2pt}
While generating all step-wise images in a single denoising pass enhances stylistic and background consistency, it struggles with recipes involving numerous small ingredients—especially in \textit{stir-fry dishes} where chopped ingredients are scattered and repeatedly appear in small visual regions. In such cases, simultaneous denoising fails to preserve crucial visual attributes such as shape, color, and texture, and may even omit ingredients. We refer to this as the \textbf{Tiny Ingredient Continuity Problem} (see Fig.~\ref{fig:CSCC}).

To address this challenge, we propose \textbf{Cross-Step Consistency Control (CSCC)}, a lightweight yet effective solution that promotes the visual continuity of fine-grained ingredients across steps, while maintaining the independence of each step’s image content. Our approach consists of two stages: 
(1) \textbf{Contextual Step Token Extraction}. 
 
We encode the entire recipe by concatenating all step instructions and feeding them into a T5 encoder, generating a unified representation. Since recurring ingredients appear in multiple steps, their token representations inherently share semantic similarities. We then segment the token sequence according to step lengths to extract context-aware tokens for each step—retaining both global context and local specificity.
Many recipe steps contain vague descriptions, such as "Pour the batter in the pan," without specifying the exact ingredients (e.g., carrot and zucchini strips). To address this, we introduce the \textbf{Cooking Agent}, based on GPT-4o \cite{achiam2023gpt}, which supplements missing ingredient details for each step. The Cooking Agent fills in any implicit ingredient information not explicitly mentioned in the recipe text, while ensuring that descriptions—such as shape and color—remain consistent across steps. This consistency is crucial for maintaining ingredient coherence across images, enabling greater consistency  in Contextual Step Token Extraction.
As shown in Fig.\ref{fig:CSCC} and Supplementary A.3, these step-specific tokens effectively capture ingredient-level consistency, allowing for coherent representation of shared ingredients across steps.
(2) \textbf{Context-Aware Fusion via Weighted Averaging}. 
To reinforce ingredient continuity across steps, we combine the globally informed step tokens with the tokens generated from individually encoding each step from SRC. This combination is achieved through a weighted averaging approach, which strikes a balance between maintaining the unique details of each step and ensuring consistency in ingredient appearance. As a result, the model preserves fine-grained details—such as the color and shape of ingredients—across all generated images, while keeping the independence of each step's content intact. The process can be represented as:
\begin{equation}
    \dot{C}^{(n)}[0:t^{(n)}]=C^{(n)}[0:t^{(n)}] + \lambda * C^{recipe}[b^{(n)}:b^{(n)}+t^{(n)}],
    \label{cscc}
\end{equation}
where $C^{(n)}$ represents the tokens obtained by individually decoding the $n$-th step and $C^{recipe}$ represents the tokens obtained by decoding the entire recipe together. $b^{(n)}$ denotes the starting position of the $n$-th step in the token sequence $C^{recipe}$. $t^{(n)}$ represents the length of the $n$-th step. $\lambda$ is a weight factor that balances the contribution of the recipe-wide context and the individual step decoding.
\section{Experiment}

\subsection{Experiment Settings}

\textbf{Datasets.} We conduct experiments on RecipeGen~\cite{zhang2025recipegen} and VGSI~\cite{zhou2018towards}. Details on how the two datasets are used can be found in Supplementary A.1.

\textbf{Implementation Details.} We evaluate our model in both training-free and training-based settings. Detail is in Supplementary A.2.

\textbf{Evaluation metrics.} 
We evaluate CookAnything on two datasets: RecipeGen~\cite{zhang2025recipegen} and VGSI-Recipe~\cite{yang2021visual}, using the following metrics in Tab.\ref{tab:evaluation-metrics}.

\begin{table}[h]
\centering
\small 
\renewcommand{\arraystretch}{1.5}
\caption{Evaluation metrics for CookAnything.}
\begin{tabularx}{\linewidth}{p{2cm} X}
\toprule
\textbf{Metric} & \textbf{Description} \\
\midrule
\textbf{Step \newline Flexibility} & Indicates whether the model can generate a variable number of step images in a single pass. \\
\textbf{Joint \newline Generation} & Indicates whether the model can generate all step images simultaneously to ensure consistency and efficiency. \\
\textbf{Goal \newline Faithfulness} & CLIP~\cite{radford2021learning} similarity between the final image and last-step caption, measuring alignment with the overall goal. \\
\textbf{Step \newline Faithfulness} & Assesses each image’s alignment with its step caption using CLIP and contextual consistency with the recipe via GPT-4o~\cite{achiam2023gpt}. Detailed prompt can be found in the Supplementary A.4.\\
\textbf{Cross-Step \newline Consistency} & Based on StackedDiffusion~\cite{menon2024generating} and DINOv2~\cite{dino}, uses $l_2$ distance and step count difference to assess visual and numerical consistency. \\
\textbf{Ingredient\newline  Accuracy} & Uses GPT-4o and manual inspection to verify whether the expected ingredients are visually present in each step and to check for any omissions. Detailed prompt is in Supplementary A.4.\\
\textbf{Usability} & Evaluates spatial alignment in jointly generated images to avoid layout collapse or misalignment, using GPT-4o and human inspection. GPT-4o scores usability across five aspects: size consistency, step clarity, content duplication, process completeness, and step count reasonableness. Full results in Tab.~\ref{tab:maintab}, details in Supplementary A.4. \\
\bottomrule
\end{tabularx}

\label{tab:evaluation-metrics}
\end{table}

\begin{figure*}[t]
  \centering
  \includegraphics[width=1\linewidth]{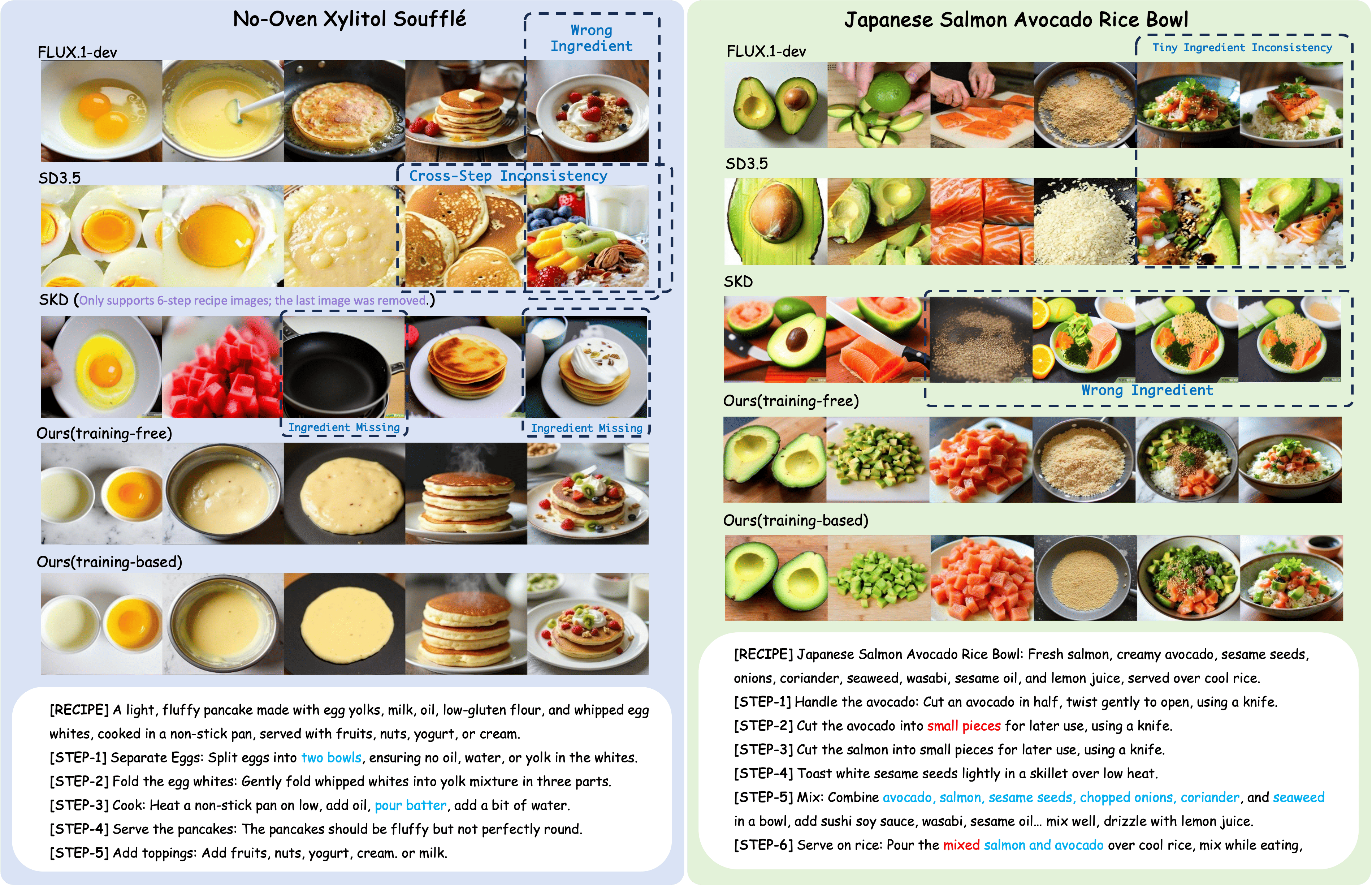} 
  \caption{Qualitative comparisons. SKD refers to StackedDiffusion, and SD3.5 refers to Stable Diffusion 3.5. Both SD3.5 Flux.1-dev and SKD exhibit issues with ingredient accuracy, discontinuous ingredient shapes, and the generation of incorrect ingredients. In contrast, our model excels in maintaining the shape and continuity of ingredients.}
  \label{fig:visualize}
\end{figure*}

\subsection{Quantitative Evaluation}

We evaluate our model against a wide range of baselines, including \textbf{UNet-based} methods: StoryDiffusion \cite{zhou2024storydiffusion}, Stable Diffusion XL (SDXL) \cite{podell2023sdxl}, and StackedDiffusion (SKD) \cite{menon2024generating}. We also compare with \textbf{DiT-based} models: Flux.1-dev , In-Context LoRA (IC-LoRA) \cite{IC-Lora}, Stable Diffusion 3.5 (SD3.5) \cite{stabilityai2024sd35large} and Regional Prompt Flux (RPF) \cite{chen2024training}. Additionally, we include \textbf{layout-aware} methods such as GLIGEN \cite{li2023gligen} and Attention Refocusing (A-R) \cite{phung2024grounded}.

As shown in Tab.~\ref{tab:maintab}, our \textbf{CookingAnything} model achieves state-of-the-art results on the RecipeGen dataset, surpassing all baselines across key metrics. It excels in Goal Faithfulness (GF) and Step Faithfulness (SF), indicating \textit{precise visual-step alignment}, and achieves the lowest Cross-Step Consistency (CSC), reflecting \textit{strong procedural coherence}. Additionally, it leads in Ingredient Accuracy (IA) and Usability (UB) under both GPT-based (G) and Human Evaluation (H), demonstrating its ability to \textit{capture fine-grained details and deliver user-preferred outputs.}

We further validate our model on on the VGSI-Recipe dataset using both \textbf{Training-Free (TF)} and \textbf{Training-Based (TB)} settings.
As shown in Tab.~\ref{tab:vgsitab}, our method consistently achieves the best performance across all metrics. These results highlight the effectiveness of our approach in producing accurate, and coherent recipe visualizations across diverse scenarios.

\begin{table*}
  \centering
  \renewcommand{\arraystretch}{1.2} 
  \tabcolsep=6pt                  
  \small  
  \caption{Comparison with Other Models in RecipeGen. SF means Step Faithfulness, IA means Ingredient Accuracy and UB means Usability. (C) means CLIP score, (G) means GPT-Score and (H) is Human evaluation. The UB metric is applicable only to methods capable of Joint Generation. TF means Traning-Free and TB means Traning-Based.}
  \begin{tabular}{@{}llcccccccccc@{}}
    \toprule
    \multirow{2}{*}{\textbf{Category}} &
    \multirow{2}{*}{\textbf{Method}} & 
    \multirow{2}{*}{\textbf{\makecell{Step\\ Flexibility}}} & 
    \multirow{2}{*}{\textbf{\makecell{Joint \\ Generation}}} & 
    \multirow{2}{*}{\textbf{\makecell{Goal \\ Faithfulness}} $\uparrow$} & 
    \multirow{2}{*}{\textbf{\makecell{Cross-Step \\Consistency}$\downarrow$} } & 
    \multicolumn{2}{c}{\textbf{SF}$\uparrow$} & 
    \multicolumn{2}{c}{\textbf{IA}$\uparrow$}  & 
    \multicolumn{2}{c}{\textbf{UB} $\uparrow$} \\
    \cmidrule(lr){7-8} \cmidrule(lr){9-10} \cmidrule(lr){11-12}
    & & & & & & (C) & (G) & (G) & (H) & (G) & (H) \\
    \midrule
    \multirow{4}{*}{UNet-based} 
      & StoryDiffusion \cite{zhou2024storydiffusion}   & \Checkmark & \XSolidBrush & 17.51& 6.01 & 25.54 & 5.30 & 7.25 & 3.41 & \textemdash{} & \textemdash{} \\
      & SDXL~\cite{podell2023sdxl}         & \Checkmark & \XSolidBrush & 27.46 & 2.98 & 29.37 & 6.79 & 7.51 & 3.71 & \textemdash{} & \textemdash{} \\
      & SKD~\cite{menon2024generating}     & \XSolidBrush & \Checkmark   & 26.62 & 0.7  & 28.53 & 4.57 & 6.67 & 2.59 & 6.43    & 3.59    \\
    \cdashline{1-12}[1pt/2pt]
    \multirow{4}{*}{DiT-based} 
      & SD3.5~\cite{stabilityai2024sd35large} & \Checkmark & \XSolidBrush & 27.42 & 2.97 & 28.77 & 6.73 & 7.58 & 3.97 & \textemdash{} & \textemdash{} \\
      & Flux.1-dev~\cite{flux1ai2024}          & \Checkmark & \XSolidBrush & 26.47 & 3.47 & 28.31 & 5.31 & 5.93 & 5.71 & \textemdash{} & \textemdash{} \\
      & IC-LoRA~\cite{IC-Lora}                  & \Checkmark & \Checkmark   & 26.07 & 9.03 & 26.58 & 4.03 & 5.50 & 3.91 & 5.34    & 4.45    \\
      & RPF~\cite{chen2024training}             & \Checkmark & \Checkmark   & 27.19 & 8.73 & 25.99 & 4.45 & 7.05 & 3.02 & 3.89    & 4.24    \\
    \cdashline{1-12}[1pt/2pt]
    \multirow{2}{*}{Layout-aware} 
      & GLIGEN~\cite{li2023gligen} & \Checkmark & \XSolidBrush & 26.99 & 2.17 & 26.72 & 5.17 & 6.16 & 5.28 & \textemdash{} & \textemdash{} \\
      & A-R~\cite{phung2024grounded} & \Checkmark & \XSolidBrush & 26.31 & 2.46 & 27.63 & 4.58 & 5.46 & 5.29 & \textemdash{} & \textemdash{} \\
    \midrule
    \addlinespace 
     \multirow{2}{*}{DiT-based} 
      & Ours (TF) & \Checkmark & \Checkmark & \underline{30.12} & \textbf{0.17} & \underline{29.80} & \underline{8.52} & \underline{9.12} & \underline{6.92} & \textbf{9.89} & \underline{7.66} \\
    
      & Ours (TB) & \Checkmark & \Checkmark & \textbf{30.59} & \underline{0.19} & \textbf{30.45} & \textbf{8.69} & \textbf{9.27} & \textbf{7.15} & \underline{9.70} & \textbf{8.48} \\
    \bottomrule
  \end{tabular}
  
  \label{tab:maintab}
\end{table*}

\begin{table}
  \centering
  \tabcolsep=6pt 
  \small         
    \caption{Comparison with Other Models in VGSI-Recipe. SF (C) and SF (F) denote CLIP-based and GPT-based Step Faithfulness, respectively.}
  \begin{tabular}{@{}lcccccc@{}}
    \toprule
    \textbf{Method} & \textbf{GF}$\uparrow$& \textbf{SF(C)}$\uparrow$ & \textbf{SF(G)}$\uparrow$
    & \textbf{CSC}$\downarrow$& \textbf{IA}$\uparrow$& \textbf{UB}$\uparrow$\\
    \midrule
    SD1.5~\cite{rombach2022high} & 27.03& 26.63& 3.41& 12.28& 5.32& \textemdash{}\\
    SD2.1~\cite{stabilityai2022sd21} & 27.03& 26.63&3.32& 11.94& 5.01& \textemdash{}\\
    SDXL ~\cite{podell2023sdxl}& 27.78& 28.32&3.60& 11.7& 5.87& \textemdash{}\\
    SD3.5 ~\cite{stabilityai2024sd35large}& 26.12& 27.03&3.43& 5.27& 5.17& \textemdash{}\\
    Flux.1-dev~\cite{flux1ai2024} & 25.82& 27.77& 2.99&5.97& 4.55& \textemdash{}\\
    IC-LoRA~\cite{IC-Lora} & 26.18& 28.25&3.77& 4.88& 5.74& 6.06\\
    SKD~\cite{menon2024generating} & 28.25& 28.26& 4.22&3.14& 6.35& 5.75\\
    RPF~\cite{chen2024training} & 28.40& 26.54& 3.25&7.12& 6.03& 3.96\\
 GLIGEN~\cite{li2023gligen} & 29.70& 29.48& 5.01&4.08& 6.93&\textemdash{} \\
 A-R~\cite{phung2024grounded} & 28.87& 28.34& 4.32&3.94& 6.31&\textemdash{} \\
    \midrule
    \addlinespace 
    \rowcolor{gray!10} 
    Ours (TF) & \textbf{31.22}&\underline{ 29.61}& \textbf{7.12}&\textbf{1.67}& \textbf{8.42}& \textbf{9.06}\\
    \rowcolor{gray!10}
    Ours (TB)& \underline{29.88}& \textbf{29.71}&\underline{6.63}& \underline{2.26}& \underline{8.07}& \underline{7.72}\\
    \bottomrule
  \end{tabular}

  \label{tab:vgsitab}
\end{table}

\subsection{Ablation Study}
Our main experiments are conducted in the Training-Based Mode, while the ablation study on the Training-Free Mode is provided in Supplementary Section A.7.

\textbf{Effectiveness of Cross-Step Consistent Control.} 
We propose the Cross-Step Consistent Control (CSCC) to ensure ingredient consistency across steps. As shown in Tab.~\ref{tab:AblationStudy}, removing CSCC leads to a notable drop in Cross-Step Consistency (CSC), demonstrating the effectiveness of our design.
We also examine the effect of the hyperparameter $\lambda$ in Equation 5, balancing the regional and context-enhanced prompts. Testing $\lambda$ values from {0, 0.2, 0.4, 0.6, 0.8, 1} (Fig. \ref{fig:longrange} (d)-(f)) reveals $\lambda = 0.2$ as optimal, so we set $\lambda = 0.2$ in this paper.

 


\begin{table}
  \centering
  \small
   \caption{Ablation Study. F-RoPE refers to the Flexible RoPE we proposed, CSCC stands for Cross-Step Consistency Control, and C-Agent refers to the Cooking Agent. SF (C) and SF (F) denote CLIP-based and GPT-based Step Faithfulness, respectively.}
  \begin{tabular}{@{}lcccc}
    \toprule
    \textbf{Method} & \textbf{GF} $\uparrow$& 
    \textbf{SF(C)}$\uparrow$ & \textbf{SF(G)}$\uparrow$& \textbf{CSC} $\downarrow$ \\ 
    \midrule
    w/o F-RoPE&28.66&29.33&7.93&\underline{0.23} \\
    w/o CSCC&\underline{30.57}&\underline{30.28}&\underline{8.67}&0.29 \\
    w/o C-Agent& 29.00& 29.59& 7.84&3.06 \\
    \midrule
    \rowcolor{gray!10}
    \textbf{Ours} &\textbf{30.59}&\textbf{30.45}&\textbf{8.69}&\textbf{0.19} \\
    \bottomrule
  \end{tabular}
 
  \label{tab:AblationStudy}
\end{table}
\setlength{\floatsep}{5pt plus 2pt minus 2pt}
\setlength{\textfloatsep}{5pt plus 2pt minus 2pt}
\setlength{\intextsep}{5pt plus 2pt minus 2pt}

\textbf{Effectiveness of Flexible RoPE.}
We evaluate Flexible RoPE by replacing it with the original RoPE. As shown in Tab.\ref{tab:AblationStudy}, removing Flexible RoPE causes a notable drop in Goal Faithfulness (1.93) and Step Faithfulness (0.98), and introduces visual inconsistencies across steps. Specifically, consecutive images exhibit blurred transitions, making step boundaries harder to distinguish (Fig.\ref{fig:rope}).


\textbf{Effectiveness of Cooking Agent.} We remove the Cooking Agent and test on diverse recipe steps, including those with vague or implicit ingredient descriptions. As shown in Tab.\ref{tab:AblationStudy}, CookingAnything still outperforms all other models in Tab.\ref{tab:maintab}, showing its ability to understand recipe text and perform well even without explicit instructions.

\textbf{Evaluation on Variable-Length Recipes.} We evaluate the performance of our model, as well as In-Context LoRA, Regional Prompting FLUX, and Flux.1-dev, on recipes ranging from 3 to 10 steps. Detailed results can be found in Supplementary Section A.7.

\subsection{Qualitative Evaluation}
Fig.\ref{fig:visualize} presents qualitative comparisons between our model and other DiT-based methods, including StackedDiffusion, Stable Diffusion 3.5 and Flux.1-dev, covering Western cuisine, and Asian dishes.  Our approach consistently demonstrates superior \textit{cross-step consistency in terms of both }\textit{global scene layout and fine-grained visual fidelity. }
For instance, it accurately maintains the shape of the soufflé and the structure of the salmon and avocado. Furthermore, our generations are more faithful to the recipe content and step-wise instructions. In contrast, other methods frequently suffer from ingredient omissions or hallucinations and show inconsistency in both background and ingredient appearance throughout the sequence.

\subsection{User Study on Perceptual Quality}

To evaluate the perceptual quality of our generated images, we conducted a user study involving 53 participants on 75 questions, assessing five key aspects: \textit{Cross-Step Consistency (CSC)}, \textit{Step Faithfulness (SF)},\textit{ Goal Faithfulness (GF)}, \textit{Aesthetic Quality (AQ)}, and \textit{Overall Appeal (OA)}. As summarized in Tab.~\ref{tab:human}, our method consistently surpasses Stable Diffusion 3.5 and Flux.1-dev across all metrics, achieving a significantly higher \textit{Aesthetic Quality} score of 60.38. These results highlight the superior visual fidelity and user preference of our approach. Further details are provided in Supplementary A.9.

\setlength{\floatsep}{5pt plus 2pt minus 2pt}
\setlength{\textfloatsep}{5pt plus 2pt minus 2pt}
\setlength{\intextsep}{5pt plus 2pt minus 2pt}
\begin{table}[h!]
  \centering
  \small
  \caption{Human Evaluation on Perceptual Quality.}
  \begin{tabular}{@{}lccccl}
    \toprule
    \textbf{Method} & \textbf{GF} $\uparrow$& 
    \textbf{SF(C)}$\uparrow$ & \textbf{CSC} $\uparrow$& \textbf{AQ }$\uparrow$&\textbf{OA}$\uparrow$\\ 
    \midrule
    SD3.5&12.08&15.60&12.70&13.58 &13.46\\
    Flux.1-dev&17.11&16.23&17.61&26.34 &16.86\\
    \midrule
    \rowcolor{gray!10}
    \textbf{Ours} &7\textbf{0.82}&\textbf{68.18}&\textbf{69.69}&\textbf{60.38} &\textbf{69.68}\\
    \bottomrule
  \end{tabular}
  
  \label{tab:human}
\end{table}



 


\section{Conclusion}

In this work, we present the \textbf{CookAnything}, a novel framework for flexible, high-fidelity illustrated recipe generation from step-wise textual instructions. By integrating \textit{Step-wise Regional Control}, \textit{Flexible RoPE}, and \textit{Cross-Step Consistency Control}, our approach addresses key limitations of prior methods, achieving accurate semantic alignment, step-wise visual disentanglement, and fine-grained ingredient continuity within a unified generation process. Extensive evaluations demonstrate that CookAnything not only produces visually coherent and semantically diverse step images, but also scales effectively to variable-length recipes under both Training-Based and Training-Free settings.


\clearpage

\section*{Acknowledgments}
This work was funded by the National Natural Science Foundation of China (Grant No. 62406126) and the Scientific Research Project of the Education Department of Jilin Province (Grant No. JJKH20250119KJ).

\bibliographystyle{ACM-Reference-Format}
\bibliography{sample-base}

\clearpage
\clearpage
\section{Supplementary Material}
Sec.\ref{sec:dataset} introduces the two datasets used in our work: RecipeGen and VGSI.
\textbf{Sec.\ref{sec:im}} provides a detailed description of our experimental settings.
\textbf{Sec.\ref{tokens}} presents experiments specifically conducted on Contextual Step Tokens, evaluated using three key metrics.
\textbf{Sec.\ref{usa}} elaborates on the definitions and usage of the Ingredient Accuracy and Usability metrics.
\textbf{Sec.\ref{cookingagent} }describes the detailed prompts used for our Cooking Agent.
\textbf{Sec.\ref{specfic}} lists the specific recipe steps corresponding to the images shown in Fig.1 of the main paper.
\textbf{Sec.\ref{moreablation}} presents additional ablation studies.
\textbf{Sec.\ref{morevis}} provides more qualitative visualization results.
\textbf{Sec.\ref{user} }describes the setup of our human evaluation study.
Finally, \textbf{Sec.~\ref{future} }discusses potential directions for future work.

\subsection{Datasets Details}
\label{sec:dataset}
In our experiments, we utilize two datasets: RecipeGen~\cite{zhang2025recipegen} and VGSI~\cite{zhou2018towards}. To the best of our knowledge, RecipeGen represents the first and currently the only large-scale dataset specifically constructed for the task of recipe image generation. It spans a wide range of cooking types and regional cuisines, including both liquid-based recipes and solid dishes. The number of steps per recipe is widely distributed (ranging from 2 to 15), which facilitates flexible multi-step image generation. The dataset consists of 21,944 recipes. For our experiments, we randomly sample 5,000 recipes from the training split for model training, and we use the official test split, comprising 4,389 recipes, for evaluation.

VGSI is a visual goal-step instruction dataset collected from WikiHow, where recipe-related samples represent only a small subset of the overall data. Compared to RecipeGen, VGSI encompasses fewer cooking types and exhibits more limited visual diversity, with a number of samples presented in a comic style. To focus on recipe-related content, we filter VGSI by the keyword “cook”, resulting in 1,157 recipes with a total of 6,417 images. We then apply an 85:15 split, randomly selecting 173 recipes as the test set. 

\subsection{Implementation Details}
\label{sec:im}
We evaluate our model under two distinct settings: training-free and training-based. In the training-based framework, individual step images are standardized to a resolution of 512×512 pixels. These images are then concatenated vertically to yield a complete multi-step visual sequence, ensuring spatial consistency across the recipe’s steps.

For the textual input, we incorporate the entire recipe text. To improve the model’s comprehension, we generate a concise recipe summary using GPT-4o, which is prepended to the full text. Moreover, each step description is explicitly marked with a prefix in the format “[step-i]”, denoting its corresponding step number. This results in a final input structure consisting of the GPT-4o-produced summary followed by all step-wise instructions.

Training is carried out on a single A100 GPU for 20,000 training steps, using a batch size of 2. We train on both the VGSI-recipe and RecipeGen datasets. Our implementation is built upon the Flux.1-dev variant, which consists of 19 DoubleStreamBlocks and 38 SingleStreamBlocks, aggregating to approximately 12 billion parameters. We apply LoRA with a rank of 16 to adapt the model, and optimization is performed using the Adam optimizer with an initial learning rate of 0.0001. We set $\alpha = 0.1$ and $\lambda=0.1$ in CookAnything.

\subsection{Experiments on Contextual Step Token}
\label{tokens}

In CookAnything, we concatenate all step instructions into a single sequence and pass it through the T5 encoder to obtain a unified token representation for the entire recipe. We then split the tokens for each step. In this section, we visualize the tokens representing each step and test the Goal and Step Faithfulness, as well as Cross-Step Consistency. This approach, where we only visualize the step tokens after the entire decoding process, results in performance metrics that outperform all models listed in the Tab.\ref{tab:contextual}, except for CookAnything.

\begin{table}
  \centering
  \small
   \caption{Experiments in Contextual Tokens.}
  \begin{tabular}{ccc}
    \toprule
     \textbf{\makecell{Goal \\ Faithfulness}} $\uparrow$& 
    \textbf{\makecell{Step \\ Faithfulness}}$\uparrow$& \textbf{\makecell{Cross-Step \\Consistency}} $\downarrow$\\
 \midrule
 
    29.91&28.70&1.8\\
    \bottomrule
  \end{tabular}
  \label{tab:contextual}
\end{table}

\subsection{More details about Step Faithfulness, Ingredient Accuracy, and Usability}

\begin{table}
  \centering
  \small
  \caption{Detailed result in Usability.}
  \begin{tabular}{@{}lccccc} 
    \toprule
    \textbf{Method} & \textbf{\makecell{ISC}}$\uparrow$ & 
    \textbf{\makecell{CSR}}$\uparrow$
    & \textbf{\makecell{DIC}}$\uparrow$  
    & \textbf{\makecell{PCL}}$\uparrow$
    & \textbf{\makecell{RNS}}$\uparrow$ 
    \\ 
    \midrule
    IC-LoRA&1.98&0.86&0.92& 0.79&0.78\\ 
    SKD&\textbf{2.00}&1.07&1.29& 1.04&1.03\\ 
 
    RPF&1.75&0.52&0.63& 0.50&0.50\\ 
    \midrule
    \addlinespace
    \rowcolor{gray!10}
 Ours (TF)& 1.99& \textbf{1.98}& \textbf{1.97}& \textbf{1.98}&\textbf{1.97}\\ 
 \rowcolor{gray!10}
 Ours (TB)& 1.97& 1.96& 1.85& 1.96&1.96\\ 
 \bottomrule
  \end{tabular}
  \label{tab:xxx}
\end{table}

In addition to using CLIP to compute \textbf{Step Faithfulness}, we also employ \textbf{GPT-4o} to assess the semantic alignment between the generated image and the original recipe text. Specifically, we design a five-level scoring system to evaluate whether the \textbf{ingredient shapes}, \textbf{containers}, and \textbf{states} depicted in the image accurately correspond to those described in the respective recipe step. The prompt for GPT-4o is in Fig.\ref{fig:stepfa}.
.
\textbf{Ingredient Accuracy} evaluates whether each step includes the correct and relevant ingredients. Existing models often suffer from \textbf{ingredient omission, confusion, or hallucination}, especially in multi-step cooking processes where ingredients appear, transform, or disappear over time. To address this, we assess both the presence and correctness of visible ingredients at each step, using a combination of GPT-based scoring and human verification.  The prompt used for Ingredient Accuracy is in Fig.\ref{fig:ingredientacc}.

As shown in Fig.\ref{fig:usb}, we evaluate the Usability  based on five key aspects. First, \textbf{Image Size Consistency (ISC)} ensures that the dimensions of all sub-images are uniform, with no issues such as incorrect cropping or inconsistent sizes that could hinder understanding. Second, \textbf{Clarity of Step Representation (CSR)} assesses whether each sub-image clearly represents a distinct cooking step and aligns with a specific step in the recipe, making it easy for users to follow the process. Third, \textbf{Duplication of Image Content (DIC)} checks for repetition in image content, such as identical perspectives or similar compositions, ensuring that the images remain diverse and informative. Fourth, \textbf{Process Completeness and Logic (PCL)} evaluates whether the image sequence accurately shows the entire recipe process—from preparation to the finished product—and whether the sequence logically matches the recipe text. Finally, \textbf{Reasonableness of Number of Steps (RNS)} verifies whether the number of sub-images aligns with the number of steps in the recipe, ensuring the image sequence is neither too sparse nor overloaded. These criteria together ensure that the generated images effectively represent the cooking process, are logically structured, and provide clear guidance for the user. The detailed result is in Tab.\ref{tab:xxx}. CookAnything performs the best on CSR, DIC, PCL, and RNS. However, the SKD model performs well in \textbf{Image Size Consistency} because it is fixed to output only six images, which allows for consistent cropping and uniform image sizes. Despite this, its output does not align with the number of steps in the text.

\textbf{Details about Human Evaluation in  Ingredient Accuracy, and Usability.}
\label{usa}
We conducted an evaluation on RecipeGen, where we selected 780 recipes and over 5000 images from 12 models for assessing ingredient accuracy. These were rated by 13 evaluators. Additionally, for usability, we chose 1300 recipes from four models to conduct a comprehensive review. The evaluation focused on various aspects of usability, such as the clarity of step representation, image consistency, and the logical flow of the recipe process. This thorough testing across both ingredient accuracy and usability helps ensure that our models not only generate correct ingredient representations but also provide a seamless user experience in terms of clarity and coherence.
\subsection{Prompt for Cooking Agent}
\label{cookingagent}
We employ GPT-4o as our \textit{Cooking Agent} to generate step-wise visual descriptions of recipe images. To ensure consistency across steps, we design two specialized prompts that guide the language model to produce structured and detail-rich outputs. A key objective of this design is to \textbf{maintain descriptive consistency}: once an ingredient is described in a particular form—e.g., “thinly sliced cucumber” in Step 1—the same terminology is preserved in subsequent steps unless its physical state has been explicitly altered (e.g., stir-fried, softened, browned). This consistency is critical for our \textbf{Cross-Step Consistency Control (CSCC)} block, which aligns visual representations across multiple image. The prompt is in Fig.\ref{fig:cookingagent}

\subsection{Specific Recipe for Fig.1 and 10 in Main Paper}
There are prompts for examples in Fig.1 of main paper:
\label{specfic}
\textbf{Example 1}: <Goal>: Broccoli Egg Salad. <Step-1> Blanching broccoli: Boil water in a pot, add some salt and cooking oil, and blanch the cleaned broccoli until cooked, then remove and set aside. 
<Step-2>  Preparing ingredients: Add the cut pieces of boiled eggs. 
<Step-3>  Final dish: Finished dish. 

\textbf{Example 2}: <Goal>: Hamburger. <Step-1>: Knead the dough: Mix all the ingredients except the butter and sesame seeds, and slowly add warm water while stirring with chopsticks into a flocculent state, then knead by hand to form a dough. Once it reaches the initial expansion stage, add the butter and continue kneading until the dough forms a glove film. <Step-2> Treat the dough: Deflate the fermented dough, evenly divide it into 4 portions, cover with cling film, and let rest for 10 minutes. Round the rested dough, brush with water, coat with sesame seeds, and let it ferment in the oven for another 30 minutes. 
<Step-3> Bake the dough: Place the fermented dough into a preheated oven at 160°C upper tube and 140°C lower tube, bake for about 16 minutes until done. The burger buns are now ready. 
<Step-4> Pan-fry patties and eggs: Shape the mixed chicken into patties, pan-fry until golden brown on both sides, then fry the eggs until done and set aside. 
<Step-5> Assemble the burger: Slice the burger bun horizontally in two, place a lettuce leaf and a chicken patty on one half, squeeze a little ketchup on the patty, then add cheese, egg, bell pepper rings, and top with the other bun half. 

\textbf{Example 3}: <Goal>: Mexican Chicken Wrap. <Step-1>: Prepare the tortilla: I used semi-finished tortillas, slightly fry them or microwave for one minute. <Step-2>:  Prepare the vegetables: Get lettuce and carrots ready.  <Step-3>: Fry chicken breast: Fry until golden brown (double fry for a crispier texture). <Step-4> Lay out tortilla: Place lettuce leaves and carrots on one side of the tortilla. <Step-5>Add chicken breast: Add the fried chicken breast to the tortilla. <Step-6>  Roll the tortilla: Roll up the tortilla. Cut the tortilla in half, and the Mexican chicken wrap is ready.

\textbf{Example 4}: <Goal>: Tomato and Scrambled Egg Rice. <Step-1>  Prepare ingredients: Rinse the rice and cook it in a rice cooker; beat eggs with a pinch of salt. <Step-2> Cook the eggs: Heat oil in a pan, pour in the beaten eggs, quickly stir to scramble, and set aside. <Step-3> Handle tomatoes: Make cross cuts on tomatoes, blanch in hot water to remove the skins, and cut into pieces. <Step-4> Sauté garlic and scallions: Heat oil in a pan, sauté garlic slices and white part of scallions on low heat until fragrant. 
<Step-5> Cook tomatoes: Add tomatoes, stir-fry until juicy, then add sugar and salt, and stir evenly. 
<Step-6> Combine ingredients and reduce the sauce: Add the scrambled eggs and cook until the sauce thickens, adding a bit of water if necessary. 
<Step-7>  Finish the dish: Pour the cooked tomato and eggs mixture over the steamed rice, and sprinkle with chopped scallions.

\textbf{Example 5}: <Goal>: Dried Fruit Pound Cake. <Step-1> Prepare ingredients: Prepare the ingredients. 
<Step-2>  Dried fruits: Process and cut the dried fruits into small pieces. 
<Step-3>  Melt and mix butter: Melt butter over water bath, beat with a whisk at low speed until smooth. Add salt and powdered sugar, mix briefly, then beat at low speed until combined. 
<Step-4>  Add egg mixture: Beat the eggs. Gradually add the beaten eggs to the butter in three portions, mixing well at low speed each time. 
<Step-5>  Mix flour and dried fruits: Sift in the mixture of low-gluten flour, baking powder, and almond flour. Fold with a spatula until it’s smooth. Then fold in the dried fruits. 
<Step-6>  Prepare for baking: Line the mold with parchment paper, and preheat the oven to 135°C. Pour the cake batter into the mold, smooth the surface, and tap the mold lightly to remove air bubbles. 
<Step-7>  Bake the cake: Place the mold in the oven at 135°C on the middle-lower rack for 30 minutes. Insert a toothpick into the cake; if it comes out clean, the cake is done. 
<Step-8>  Cool and cut: Take the cake out of the oven, with a beautiful golden color. Let it cool slightly and then cut into pieces. 

There are example for fig.10 of main paper:
 
\textbf{Example (a)}: <Goal>: How to Eat Kimchi.
<Step-1> Eat kimchi out of the jar for an effortless snack.<Step-2> Serve individual pieces of kimchi with toothpicks to easily share it. <Step-3> You can eat kimchi straight out of the fridge, or you can throw it in a small skillet and heat it up with 1 US tbsp (15 mL) of vegetable oil.

\textbf{Example (b)}: <Goal>: How to Use Emu Oil for Health and Skin Benefits. <Step-1> Place a small dab of the oil on the palm of your hand or on the affected area and rub it in until its clear. Within a short amount of time, you should feel relief and notice swelling go down. You may apply emu oil when your back or neck feels swollen or sore. Emu oil can be purchased online or in your local pharmacy. Use the oil once or twice a day. <Step-2> Emu oil has a slight pain-killing effect when applied to the skin, so rub it onto a scrape or bruise to reduce your pain once a day. The antioxidants found in the oil can also help prevent additional damage or further infection. Seek medical help if you have large, deep cuts. <Step-3> Gently rub the area with the emu oil once a day until its completely absorbed by your skin. The oil will reach deep into your skin and alleviate the pain quickly while the sunburn heals. Speak with your doctor or dermatologist to determine if emu oil is a good option for you. Have a friend help you get the oil on hard-to-reach areas such as your back. You can also use emu oil as a natural sunscreen. Apply the oil as you would with a regular sunscreen. 

\textbf{Example (c)}: <Goal>:How can I repair peeling for a faux leather sofa with a vinyl repair kit?
<Step-1>: Mix paint colors until they match the sofa.
<Step-2>: Brush paint onto the affected area.
<Step-3>: Apply texture relief paper to the paint, if desired.
<Step-4>: Use a heat tool with the paper for 2 minutes and finish.The sofa looks as good as new.

\subsection{More Ablation Study}

\label{moreablation}
Table \ref{tab:AblationStudy2} presents the ablation results of our Training-Free module under different levels of contextual integration, controlled by the hyperparameter $\lambda$. As $\lambda$ increases from 0 to 1, we observe a gradual degradation in performance for all three metrics: Goal Faithfulness (GF), Step Faithfulness (SF), and Cross-Step Consistency (CSC). The best overall performance is achieved when $\lambda = 0.2$, which yields the highest GF and SF scores (30.12 and 29.80, respectively) and the lowest CSC value (0.17). 

\begin{table}
  \centering
 \caption{Effect of $\lambda$ on Generation Performance.}
  \begin{tabular}{@{}lccc}
    \toprule
    $\lambda$& \textbf{\makecell{Goal\\Faithfulness}} $\uparrow$& 
    \textbf{\makecell{Step\\Faithfulness}}$\uparrow$& \textbf{\makecell{Cross-Step\\ Consistency}} $\downarrow$ \\ 
    \midrule
    0&29.99&\underline{29.79}&\underline{0.19}\\
 0.2& 
\textbf{30.12}& 
\textbf{29.80}&
\textbf{0.17}\\
 0.4& 

\underline{29.99}& 
29.70&
0.3\\
 0.6& 

29.78& 
29.43&
0.81\\
 0.8& 

29.39& 
29.01&
1.66\\
 1.0& 

29.18& 
28.87&
3.15\\
 \bottomrule
  \end{tabular}
  \label{tab:AblationStudy2}
\end{table}

\subsection{More Visualization Result}
\label{morevis}
\begin{figure*}
    \centering
    \includegraphics[width=1\linewidth]{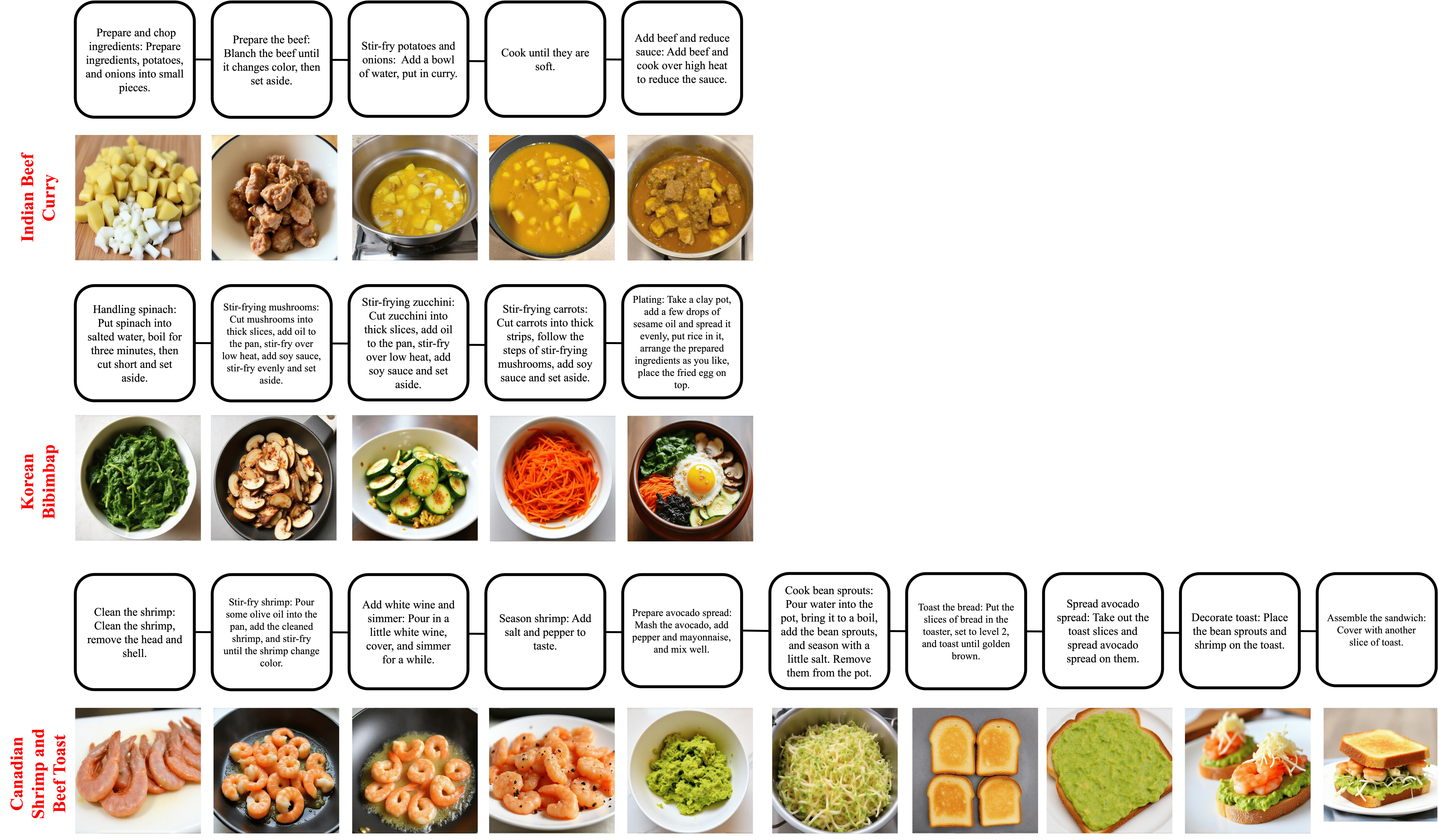}
    \caption{Visualization of dishes from different regions.}
    \label{fig:vis4}
\end{figure*}

\begin{figure*}
    \centering
    \includegraphics[width=1\linewidth]{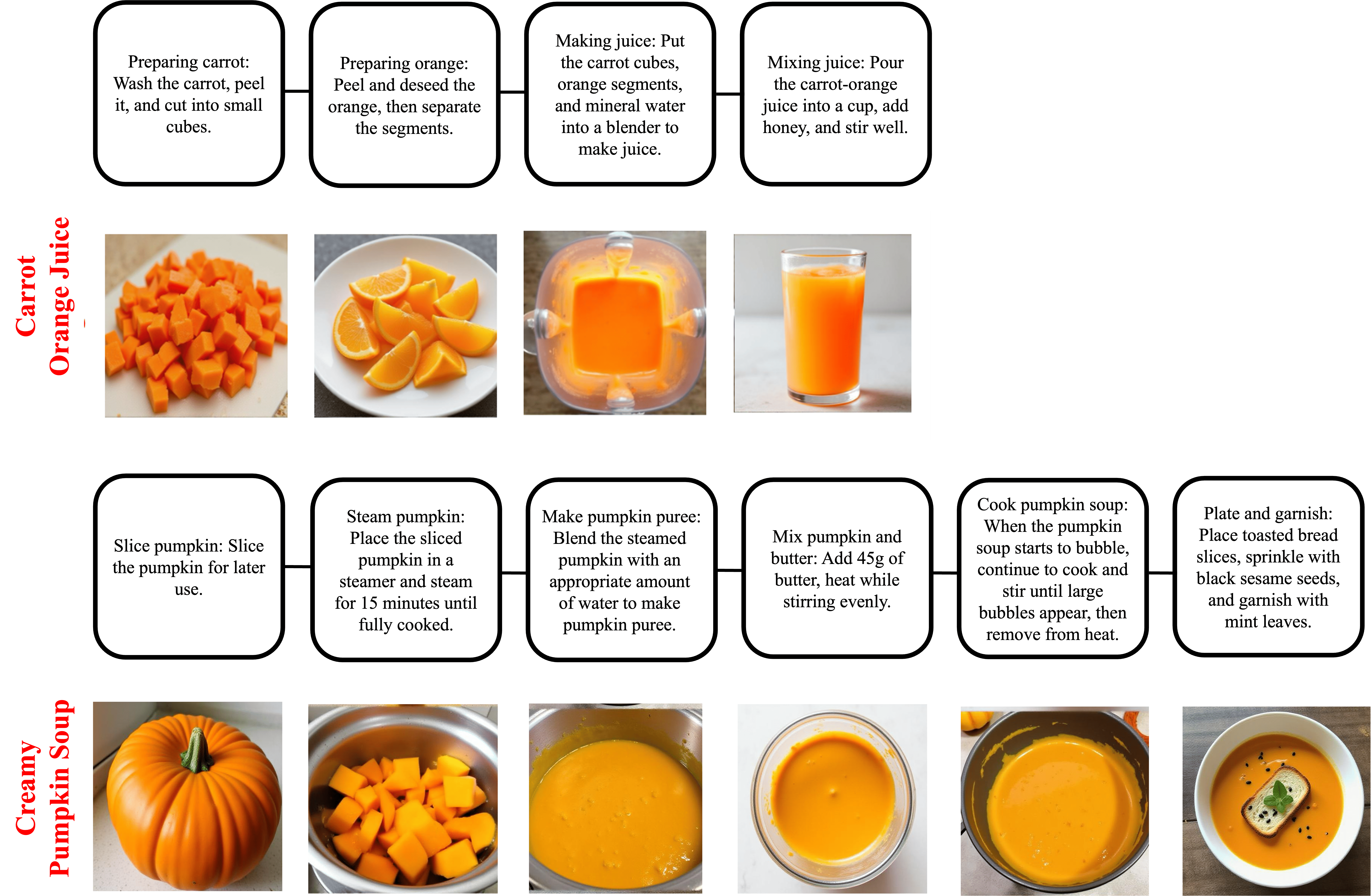}
    \caption{Visualization about liquid.}
    \label{fig:vis1}
\end{figure*}

\begin{figure*}
    \centering
    \includegraphics[width=1\linewidth]{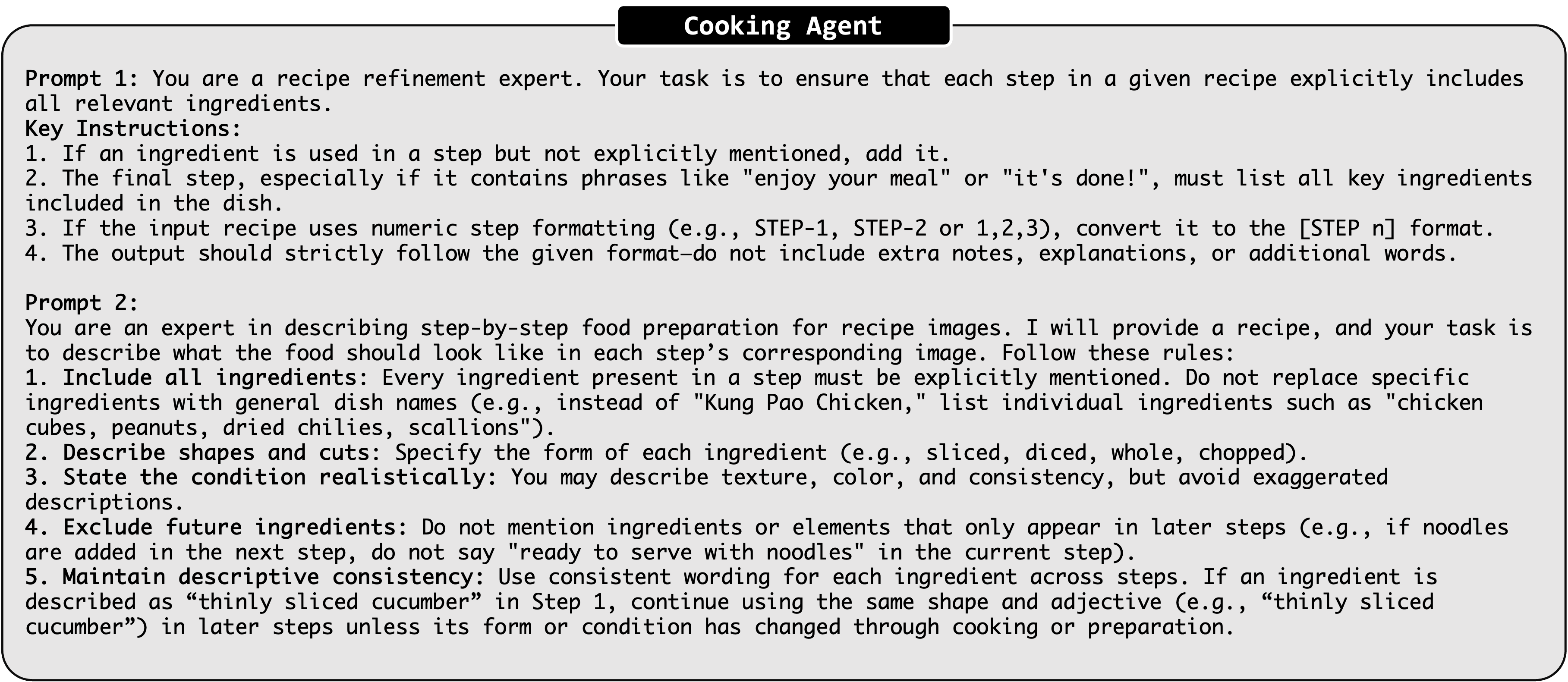}
    \caption{Prompt for refining recipe caption, with GPT-4o}
    \label{fig:cookingagent}
\end{figure*}

\begin{figure*}
    \centering
    \includegraphics[width=1\linewidth]{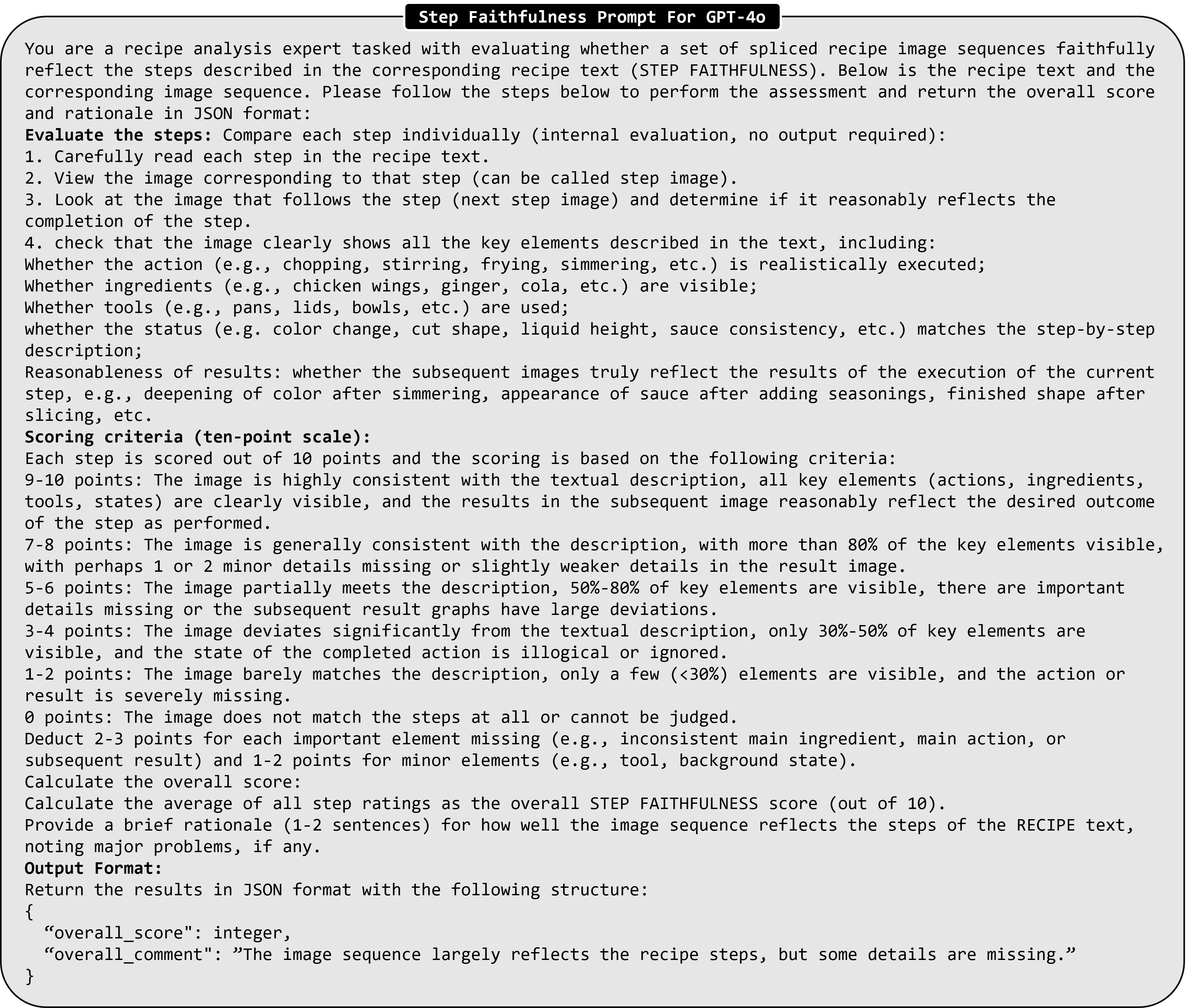}
    \caption{Prompt to measure the Step Faithfulness of the generated image, with GPT-4o}
    \label{fig:stepfa}
\end{figure*}

\begin{figure*}
    \centering
    \includegraphics[width=1\linewidth]{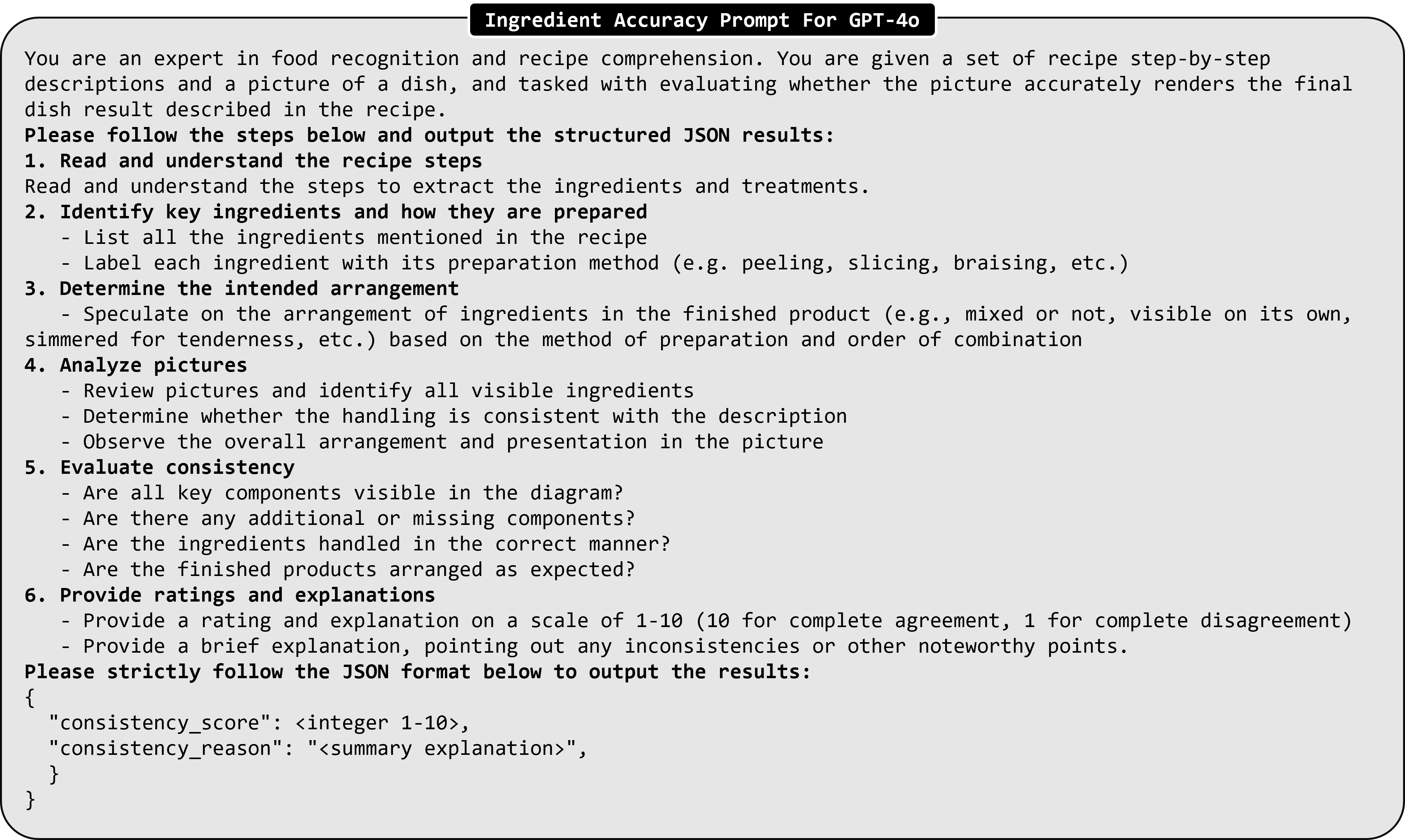}
    \caption{Prompt to measure the Ingredient Accuracy of the generated image, with GPT-4o}
    \label{fig:ingredientacc}
\end{figure*}

\begin{figure*}
    \centering
    \includegraphics[width=1\linewidth]{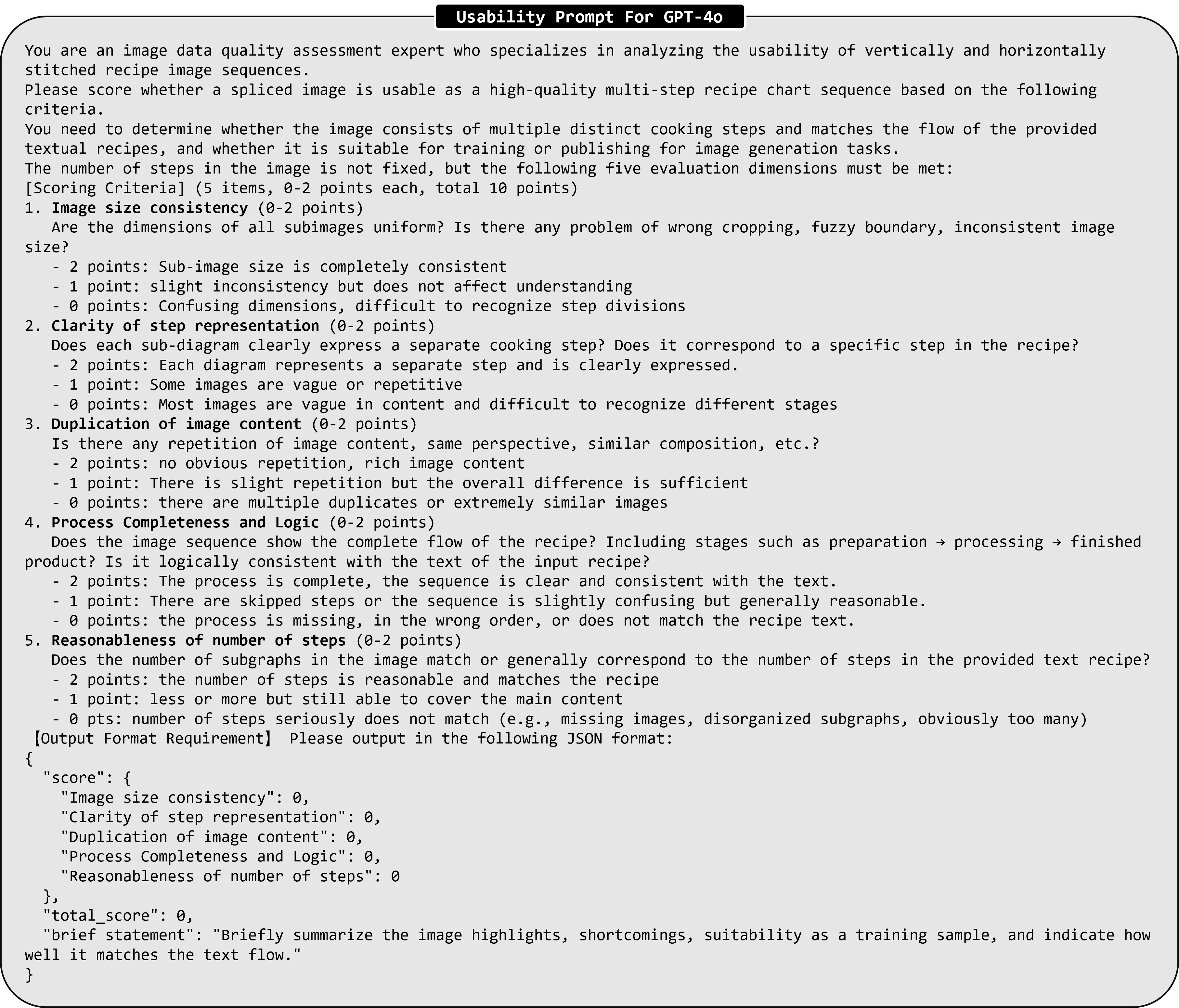}
    \caption{Prompt to measure the Usability of the generated image, with GPT-4o}
    \label{fig:usb}
\end{figure*}

In this section, we visualize results across three major food categories. First, beyond the Asian and Western dishes already shown in Fig.1, 3, 4, and 6 of the main paper, we present additional examples of \textbf{diverse regional cuisines }in Fig.\ref{fig:vis4}, demonstrating the model’s adaptability to various cultural food styles. Second, we showcase our model's performance on \textbf{liquid-based dishes} in Fig.\ref{fig:vis1}, which often pose challenges due to their fluid textures and fine-grained visual details.


\begin{figure*}[t]
  \centering
 
  \includegraphics[width=1\linewidth]{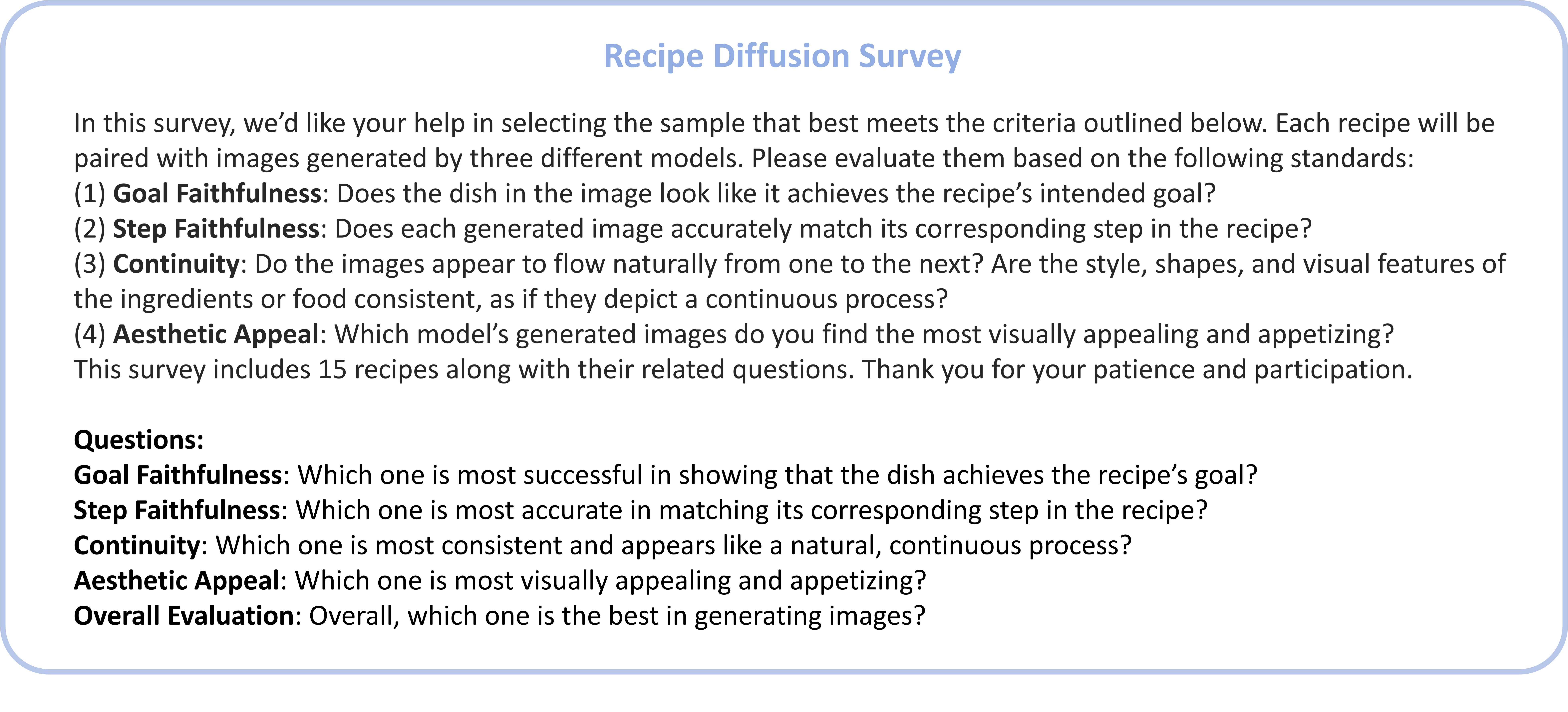} 

  \caption{Template for Human Study.}
  \label{fig:humanstudy}
\end{figure*}
\subsection{User Study}
\label{user}
The question for User Study is in Fig.\ref{fig:humanstudy}.
\subsection{Future Work}
\label{future}
Beyond cooking, our method lays a foundation for structured visual generation in broader procedural domains such as instructional manuals, scientific workflows, and educational storytelling. Future work will explore extending our framework to multimodal video generation, interactive editing, and alignment with real-world cooking data.

\end{document}